\documentclass[cameraready]{Interspeech}

\title{TTS-PRISM: A Perceptual Reasoning and Interpretable Speech Model for Fine-Grained Diagnosis}

\author[affiliation={1}]{Xi}{Wang}
\author[affiliation={2}]{Jie}{Wang}
\author[affiliation={2}]{Xingchen}{Song}
\author[affiliation={1}]{Baijun}{Song}
\author[affiliation={1}]{Jingran}{Xie}
\author[affiliation={3}]{Jiahe}{Shao}
\author[affiliation={1}]{Zijian}{Lin}
\author[affiliation={2}]{Di}{Wu}
\author[affiliation={2}]{Meng}{Meng}
\author[affiliation={2}]{Jian}{Luan}

\author[affiliation={1}, correspondingauthor]{Zhiyong}{Wu}

\address{
$^1$Tsinghua University, China \\
$^2$MiLM Plus, Xiaomi Inc., China \\
$^3$The University of Tokyo, Japan
}

\email{xi-wang24@mails.tsinghua.edu.cn, zywu@sz.tsinghua.edu.cn}

\keywords{speech quality assessment, automatic evaluation, Mandarin Chinese, speech objective evaluation}

\usepackage{comment}
\usepackage{graphicx}
\usepackage{subcaption} 
\usepackage{amsmath}
\usepackage{microtype} 
\usepackage{float}
\usepackage{array}
\usepackage{booktabs}
\usepackage{multirow}

\begin{document}

\maketitle

\begin{abstract}
While generative text-to-speech (TTS) models approach human-level quality, monolithic metrics fail to diagnose fine-grained acoustic artifacts or explain perceptual collapse. To address this, we propose TTS-PRISM, a multi-dimensional diagnostic framework for Mandarin. First, we establish a 12-dimensional schema spanning stability to advanced expressiveness. Second, we design a targeted synthesis pipeline with adversarial perturbations and expert anchors to build a high-quality diagnostic dataset. Third, schema-driven instruction tuning embeds explicit scoring criteria and reasoning into an efficient end-to-end model. Experiments on a 1,600-sample Gold Test Set show TTS-PRISM outperforms generalist models in human alignment. Profiling six TTS paradigms establishes intuitive diagnostic flags that reveal fine-grained capability differences. TTS-PRISM is open-source, with code and checkpoints at \url{https://github.com/xiaomi-research/tts-prism}.
\end{abstract}

\section{Introduction}

Driven by the rapid evolution of large-scale generative models, modern Text-to-Speech (TTS)~\cite{du2025cosyvoice, chen2025f5, wang2024maskgct,hu2026qwen3,zhou2025indextts2,xie2025fireredtts} systems have achieved human-level capabilities. However, the traditional Mean Opinion Score (MOS)~\cite{streijl2016mean} faces a ``black box'' dilemma: its single scalar obscures real capabilities in pronunciation, prosody, and emotion, and fails to capture subtle artifacts that cause perceptual collapse. This forces the evaluation paradigm to shift from holistic scoring to precise diagnosis.

Existing paradigms, however, struggle to meet this precise diagnostic demand. First, global scalar and preference-driven paradigms~\cite{saeki2024speechbertscore,zhang2024speechalign,gao2025emo,zhang2025speechjudge,ji2025wavreward} capture holistic naturalness. Yet, their sentence-level aggregation dilutes sensitivity to localized acoustic artifacts. Second, recent works improve interpretability via multi-dimensional scores and textual explanations~\cite{manakul2025audiojudge,wang2025speechllm,chen2025read,zhan2025vstyle}. However, their schemas primarily target high-level perception (e.g., artistic expression), ignoring fine-grained acoustic details and language-specific phonetics. Furthermore, the absence of explicit scoring criteria yields formulaic rationales, failing to provide actionable diagnostic feedback.

To address these challenges, we propose TTS-PRISM, a fine-grained multi-dimensional diagnostic framework. First, to establish objective anchors for ambiguous perceptual evaluations, we construct a hierarchical evaluation schema~\cite{bai2024mt}. Through explicit quantitative scoring criteria, we map subjective assessments into 12 complementary dimensions~\cite{wang2025audio}, as illustrated in Figure 1. Second, we design a targeted data synthesis pipeline, incorporating adversarial perturbations and expert anchors to sharpen discriminative capability on samples from the long-tail. Finally, we devise a schema-driven instruction tuning strategy. Grounded in comprehensive scoring criteria, it enables the model to balance a global perspective with the acute detection of fine-grained acoustic flaws.



Our main contributions are:
\noindent\textbf{(1) Fine-grained Mandarin Speech Diagnostic Benchmark:} We establish the first multi-dimensional quantitative benchmark covering both the Basic Capability and Advanced Expressiveness layers. We formulate explicit acoustically grounded criteria for each score level across the 12 dimensions, filling the critical gap in fine-grained quantitative standards for Mandarin speech evaluation.
\noindent\textbf{(2) High-quality Diagnostic Dataset:} We construct an instruction-tuning dataset comprising 200k Mandarin samples. By incorporating real human recordings and multi-paradigm TTS synthesis, we achieve a balanced distribution of positive and negative samples and comprehensive coverage of fine-grained acoustic features.
\noindent\textbf{(3) Interpretable Diagnostic Framework:} We propose TTS-PRISM to enable precise multidimensional audio diagnosis. Extending this to system profiling, we map the unique capability distributions of leading TTS paradigms based on 12-dimensional assessments. This approach moves beyond scalar ranking to reveal specific behavioral traits and architectural tendencies. Furthermore, we open-source our complete diagnostic framework, including the explicit 12-dimensional scoring criteria, code, and model checkpoints, to facilitate future research in the community.

\begin{figure}[t]
  \centering
  \includegraphics[width=\linewidth, trim=0cm 0cm 0cm 0cm, clip]{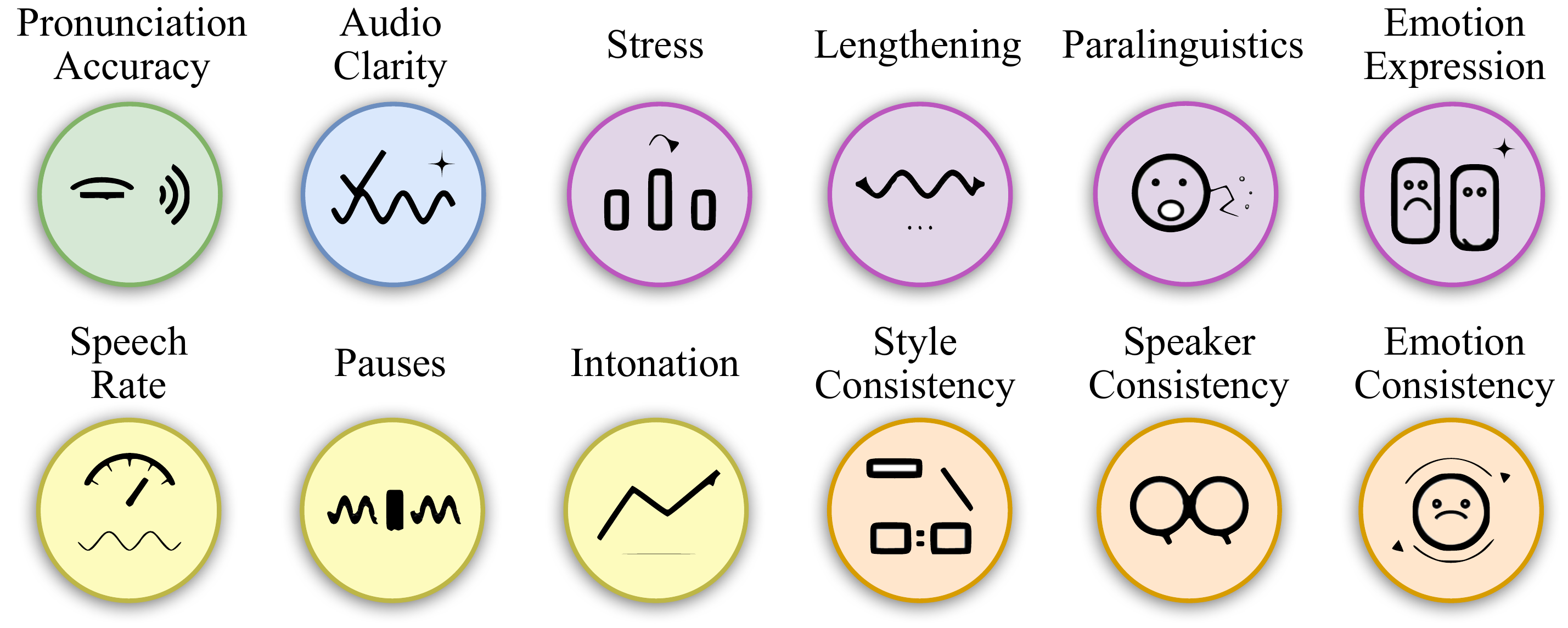}
  \vspace{-0.6cm} 
  \caption{The schema comprises 12 well-defined dimensions spanning acoustic stability and expressiveness.}
\label{fig:schema}
  \vspace{-0.6cm} 
\end{figure}

\begin{figure*}[t]
  \centering
  \begin{subfigure}[b]{0.46\textwidth}
    \centering
    \includegraphics[width=\linewidth]{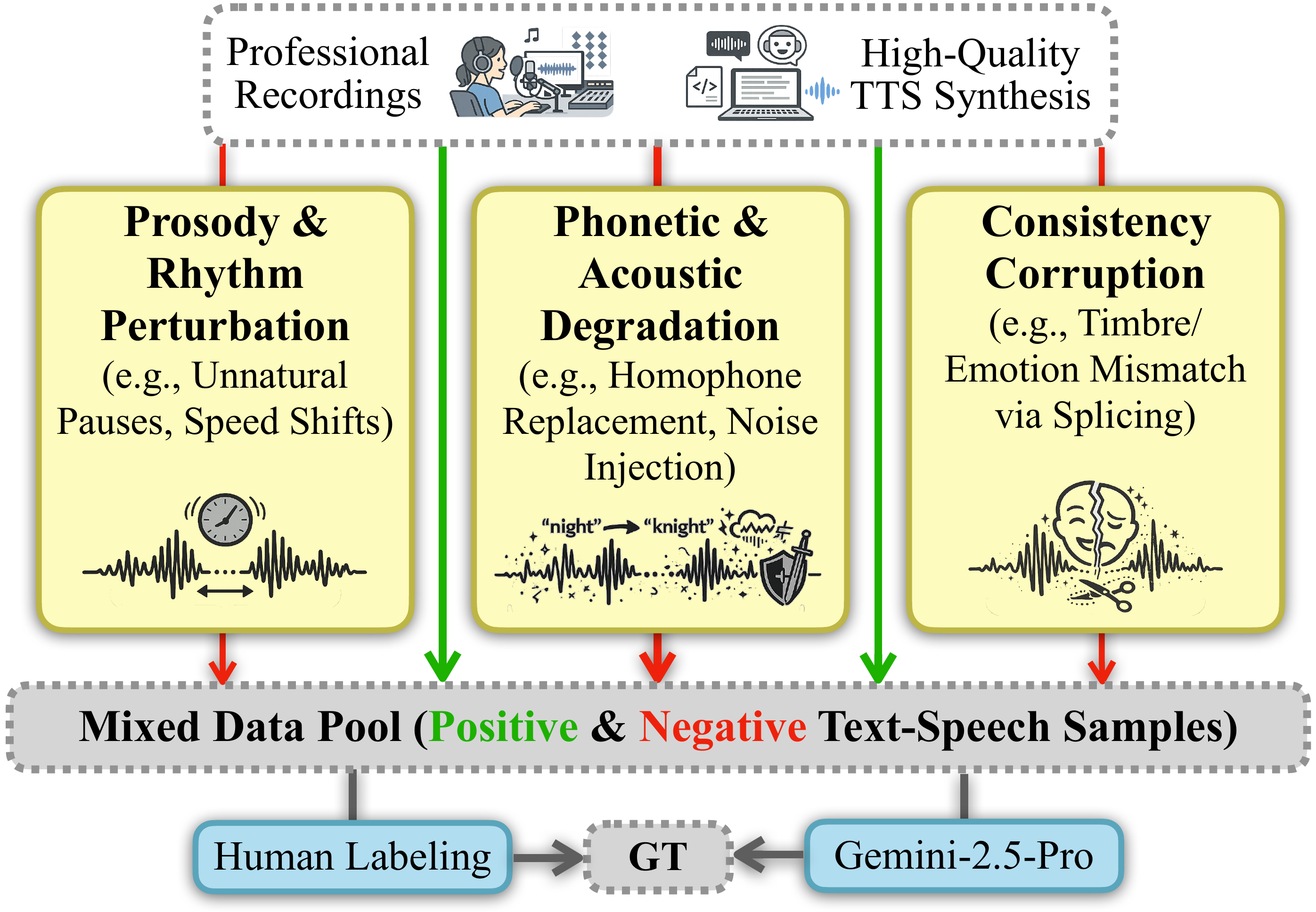} 
    \caption{Targeted Data Synthesis Strategy} 
    \label{fig:framework_a}
  \end{subfigure}
  \hfill
  \begin{subfigure}[b]{0.48\textwidth}
    \centering
    \includegraphics[width=\linewidth]{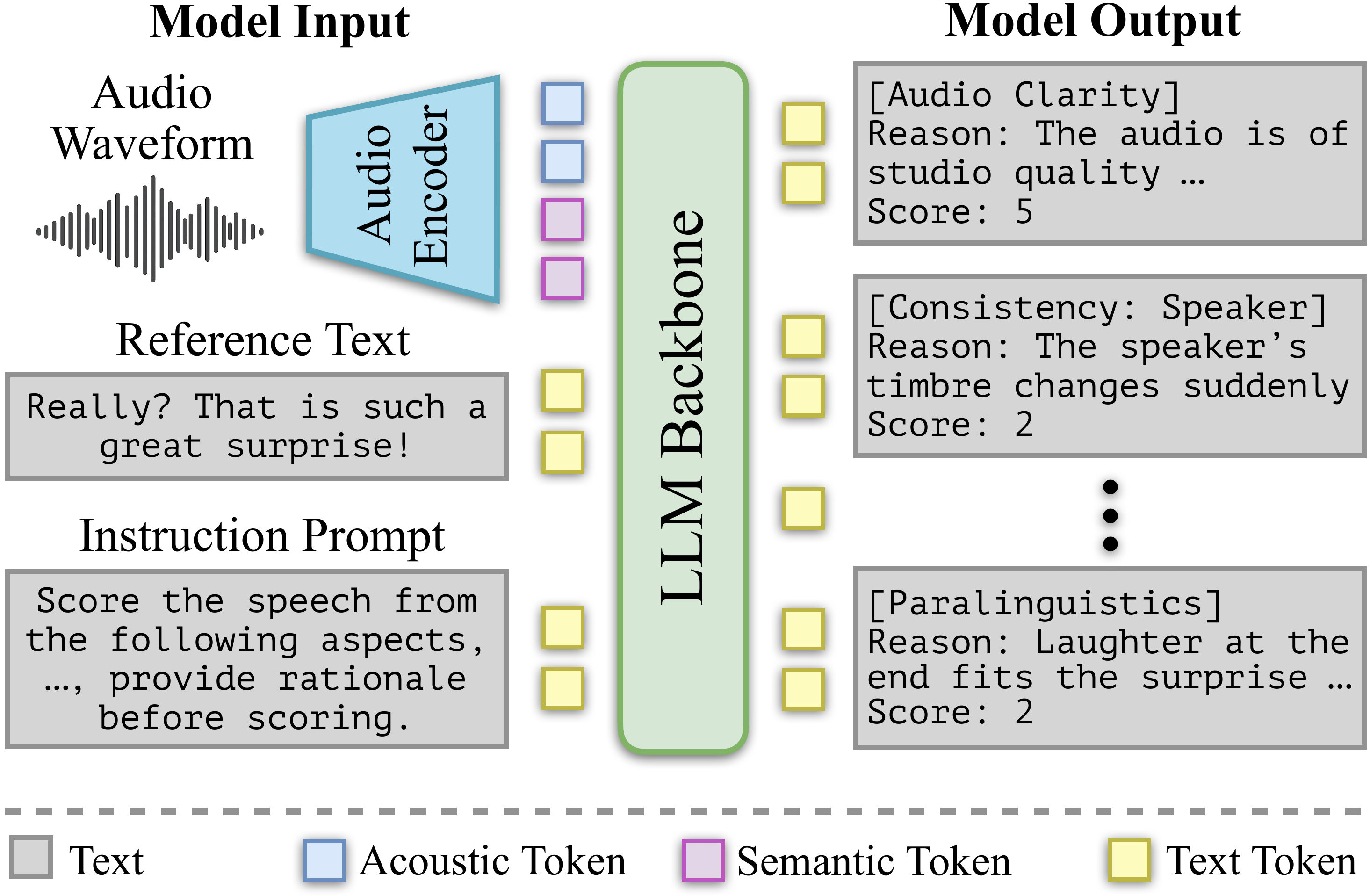}
    \caption{Diagnostic Scoring Model} 
    \label{fig:framework_b}
  \end{subfigure}
  
  \vspace{-0.2cm}
  \caption{\textbf{Overview of TTS-PRISM.} (a) The targeted synthesis strategy sharpens decision boundaries against long-tail artifacts. (b) Schema-driven instruction tuning enables 12-dimensional diagnosis via single-pass inference, balancing efficiency and interpretability.}
  \label{fig:framework}
  \vspace{-0.4cm}
\end{figure*}

\section{Methodology}

To enable fine-grained diagnosis of generative speech, we propose TTS-PRISM, a framework comprising a hierarchical evaluation schema, a targeted data synthesis pipeline, and a diagnostic scoring model. Crucially, to eliminate subjective ambiguity, we anchor each score level to explicit tolerance thresholds (e.g., defining specific artifact types permissible for a score of 4).

\subsection{Evaluation Dimensions \& Scoring Criteria}
We construct a 12-dimensional hierarchical taxonomy across 5 core domains, with 4 domains (8 sub-dimensions) forming the Basic Capability Layer and the remaining 4 sub-dimensions constituting the Advanced Expressiveness Layer.

\subsubsection{Basic Capability Layer (Score 1--5)}
This layer assesses the correctness and stability of synthesized speech, measuring whether the system meets the baseline standards for usability.

\noindent \textbf{Audio Clarity:} Evaluates physical signal quality, identifying background noise, electronic distortion, or non-target vocal residues. For instance, a Score of 4 denotes a stationary noise floor with uniform distribution and constant energy (e.g., slight Gaussian white noise or electrical hum); conversely, a Score of 2 corresponds to destructive signal distortion, including frequent popping and metallic artifacts, directly hinder intelligibility.

\noindent \textbf{Pronunciation Accuracy:} Assesses articulation correctness beyond Automatic Speech Recognition (ASR) metrics, targeting fine-grained anomalies that degrade perception: incomplete articulation, Mandarin nasal/lateral confusion (n/l), tone sandhi errors, polyphone disambiguation failures. These sub-phoneme flaws are the cause of the ``robotic'' feel in TTS models.

\noindent \textbf{Prosody Accuracy:} Encompasses three sub-dimensions: Intonation reflects syntactic structure, Pauses evaluate semantic segmentation, and Speech Rate measures rhythmic fluency.

\noindent \textbf{Consistency:} Monitors consistency of speaker identity, style, and emotional category within a single utterance.

\subsubsection{Advanced Expressiveness Layer (Score 0--2 Bonus)}
This layer captures the human-like expressive nuances of high-performance models. A Score of 0 represents ``neutral'' rather than a penalty.

\noindent \textbf{Stress:} Evaluates keyword emphasis via pitch or loudness. A Score of 2 requires significant energy concentration or pitch excursion. A Score of 1 denotes perceptible but weak emphasis, lacking sufficient acoustic prominence.

\noindent \textbf{Lengthening:} Checks whether natural syllabic lengthening occurs at phrase boundaries or emphatic points to smooth rhythm.

\noindent\textbf{Paralinguistics:} Detects non-verbal cues such as laughter, sighs, breaths, and coughs.

\noindent \textbf{Emotion Expression:} Evaluates the fullness and intensity with which the speech actualizes the sentiment inherent in the text.

\subsection{Targeted Data Construction}

Existing datasets are English-centric or exhibit positive bias~\cite{mittag2021nisqa,maniati2022somos,cooper2021voices,zhang2022wenetspeech,shi2020aishell,luo2018consistent}, blurring fine-grained decision boundaries. We therefore construct a synthesis pipeline encompassing the full quality scale, as visualized in Figure 2(a). For linguistic diversity, source texts span literary, conversational, and web corpora. On the positive side, we establish reference anchors using leading TTS paradigms and high-fidelity human speech. NVSpeech~\cite{liao2025nvspeech} and FireRedTTS-2~\cite{xie2025fireredtts} define the ceiling for paralinguistic and emotional expressions. Since Stress and Lengthening remain challenging for generative models, we use custom in-house professional recordings as gold anchors for these Advanced Expressiveness dimensions. On the negative side, we introduce perturbations in prosody and rhythm, degradations in pronunciation and audio quality, and consistency breaches. Integrating the perturbation subset from the Intelligibility Preference Speech Dataset~\cite{zhang2025advancing} further enhances sensitivity to Mandarin homophones and sub-phoneme errors.

During labeling, Gemini-2.5-Pro~\cite{comanici2025gemini} decomposes evaluation into 12 independent dimension-wise tasks, mitigating long-context instruction drift and hallucinations. We apply human-instructed rationale refinement~\cite{xu2023instructscore} to correct hallucinations in Stress and Lengthening. To address Mandarin tone sandhi and polyphones, we construct an 11k expert-annotated ``Pronunciation Gold Subset'' to inject linguistic knowledge. This yields 200k aligned samples, with source TTS~\cite{zhou2025voxcpm,du2024cosyvoice} and domain diversity visualized in Figure 3.

\begin{table*}[t]
  \centering
  \caption{Alignment between $S_{pred}$ and $S_{gt}$ on the 1,600-sample Mandarin Gold Test Set. We report LCC, SRCC, and MSE$_{\text{norm}}$ (normalized to align scales between the \textbf{Basic Capability} (1--5) and \textbf{Advanced Expressiveness} (0--2) layers).}
  \label{tab:main_results}
  \resizebox{\textwidth}{!}{%
  \begin{tabular}{l| ccc ccc ccc ccc}
    \toprule
    \multirow{2}{*}{\textbf{Dimension}} & \multicolumn{3}{c}{\textbf{Step-Audio-R1 (33B)}} & \multicolumn{3}{c}{\textbf{Qwen3-Omni (30B)}} & \multicolumn{3}{c}{\textbf{Gemini-2.5-Pro}} & \multicolumn{3}{c}{\textbf{TTS-PRISM (7B)}} \\
    \cmidrule(lr){2-4} \cmidrule(lr){5-7} \cmidrule(lr){8-10} \cmidrule(lr){11-13}
    & LCC & SRCC & MSE$_{\text{norm}}$ & LCC & SRCC & MSE$_{\text{norm}}$ & LCC & SRCC & MSE$_{\text{norm}}$ & \textbf{LCC} & \textbf{SRCC} & \textbf{MSE}$_{\text{norm}}$ \\
    \midrule
    Pronunciation Accuracy   & 0.475 & 0.423 & 0.081 & 0.169 & 0.150 & 0.202 & \textbf{0.613} & \textbf{0.530} & \textbf{0.048} & 0.511 & 0.492 & 0.073 \\
    Audio Clarity        & 0.709 & 0.690 & 0.057 & 0.665 & 0.685 & 0.065 & 0.756 & 0.594 & 0.032 & \textbf{0.815} & \textbf{0.826} & \textbf{0.018} \\
    Intonation           & 0.461 & 0.462 & 0.102 & 0.325 & 0.301 & 0.126 & \textbf{0.718} & \textbf{0.718} & \textbf{0.047} & 0.658 & 0.668 & 0.057 \\
    Pauses               & 0.541 & 0.532 & 0.068 & 0.335 & 0.374 & 0.086 & \textbf{0.731} & \textbf{0.712} & \textbf{0.043} & 0.701 & \textbf{0.712} & 0.063 \\
    Speech Rate          & 0.584 & 0.577 & 0.074 & 0.403 & 0.385 & 0.054 & 0.709 & 0.698 & 0.068 & \textbf{0.733} & \textbf{0.773} & \textbf{0.039} \\
    Speaker Consistency  & 0.591 & 0.581 & 0.064 & 0.582 & 0.563 & 0.041 & 0.733 & 0.718 & 0.029 & \textbf{0.759} & \textbf{0.752} & \textbf{0.022} \\
    Style Consistency    & 0.660 & 0.641 & 0.044 & 0.519 & 0.509 & 0.048 & 0.768 & 0.736 & 0.055 & \textbf{0.789} & \textbf{0.785} & \textbf{0.038} \\
    Emotion Consistency  & 0.657 & 0.655 & 0.073 & 0.468 & 0.452 & 0.069 & 0.752 & 0.710 & 0.040 & \textbf{0.806} & \textbf{0.794} & \textbf{0.032} \\
    \midrule
    Stress               & 0.458 & 0.413 & 0.135 & 0.313 & 0.323 & 0.072 & 0.587 & 0.551 & 0.033 & \textbf{0.648} & \textbf{0.651} & \textbf{0.027} \\
    Lengthening          & 0.416 & 0.428 & 0.224 & 0.325 & 0.324 & 0.192 & 0.558 & 0.570 & 0.115 & \textbf{0.618} & \textbf{0.620} & \textbf{0.085} \\
    Paralinguistics      & 0.541 & 0.541 & 0.069 & 0.457 & 0.460 & 0.043 & \textbf{0.751} & \textbf{0.762} & \textbf{0.017} & 0.723 & 0.737 & 0.023 \\
    Emotion Expression   & 0.707 & 0.713 & 0.067 & 0.623 & 0.615 & 0.057 & 0.808 & 0.807 & 0.080 & \textbf{0.841} & \textbf{0.838} & \textbf{0.056} \\
    \bottomrule
  \end{tabular}%
  }
\end{table*}

\subsection{Diagnostic Scoring Model}
As illustrated in Figure 2(b), we construct an end-to-end model for full-dimensional diagnosis via single-pass inference. We select MiMo-Audio~\cite{zhang2025mimo} as the backbone, utilizing its 100M-hour unsupervised pre-training for robust acoustic representations.


Building on this, we implement a schema-driven instruction tuning strategy. To mitigate hallucinations and enforce logical consistency, we construct an interleaved target sequence $Y=[R_{1},S_{1},...,R_{12},S_{12}]$ to instantiate the interpretable reasoning mechanism. Unlike the unconstrained Chain-of-Thought (CoT) typical of generalist Audio-LLMs, our rationales $R_{i}$ are strictly conditioned on explicit scoring criteria. By compelling the model to generate objective anchors $R_{i}$ before assigning scores $S_{i}$, this design acts as a crucial logical regularizer that minimizes hallucinations, as validated in Section~4.

\begin{figure}[H]
  \centering
  \includegraphics[width=\linewidth, trim=0cm 0cm 0cm 0cm, clip]{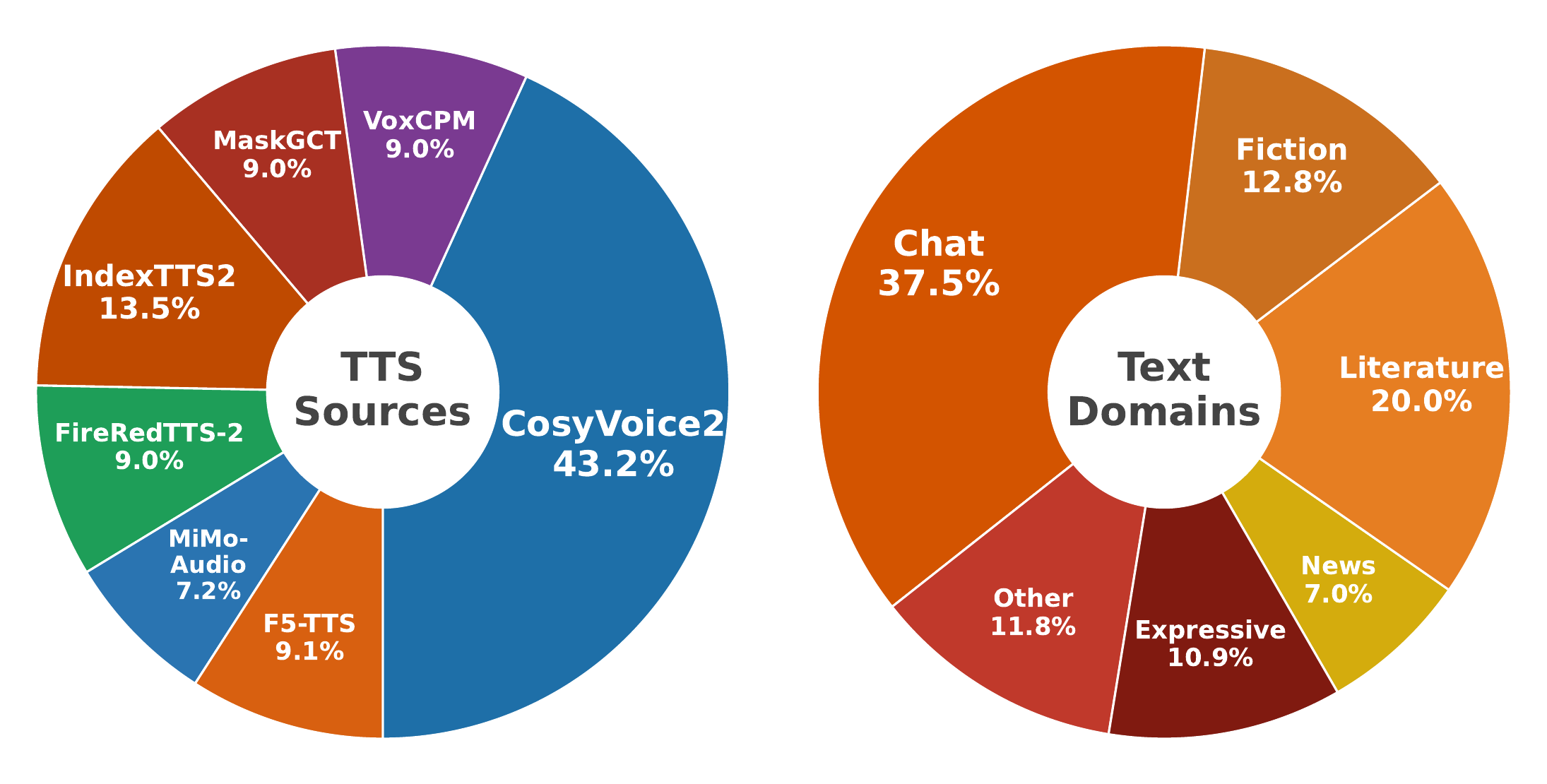}
  \vspace{-0.6cm}
  \caption{Distribution of diverse TTS sources and text domains.}
\label{fig:data_dist}
  \vspace{-0.1cm}
\end{figure}
\section{Experimental Setup}

\subsection{Dataset \& Training Configuration}
To evaluate alignment precision, we build a stratified 1,600-sample Mandarin Gold Test Set, strictly disjoint from training data, with 20\% out-of-distribution (OOD) samples (unseen TTS and real recordings) and all labels validated via consensus-based expert annotation. For training, we perform full-parameter Supervised Fine-Tuning (SFT) on MiMo-Audio with AdamW (batch size 1, fixed lr=1e-6).

\begin{table*}[t]
  \centering
  \caption{Diagnostic Profiling of leading TTS systems. Scores span the \textbf{Basic Capability} (1--5) and \textbf{Advanced Expressiveness} (0--2) layers. Based on these 12-dimensional scores, evaluators assign an intuitive \textbf{Diagnostic Flag} summarizing each system's dominant trait.}
  \label{tab:tts_profile}
  \resizebox{\textwidth}{!}{%
  \begin{tabular}{l cccccc}
    \toprule
    \textbf{Dimension} & \textbf{F5-TTS} & \textbf{CosyVoice 3} & \textbf{MaskGCT} & \textbf{Qwen3-TTS} & \textbf{FireRedTTS-2} & \textbf{IndexTTS2} \\
    \midrule
    Pronunciation Accuracy   & 4.843 & 4.850 & 4.797 & \textbf{4.860} & 4.809 & \underline{4.853} \\
    Audio Clarity            & 4.612 & \textbf{4.803} & 4.560 & \underline{4.750} & 4.580 & 4.697 \\
    Intonation               & 4.595 & \underline{4.700} & 4.550 & 4.630 & 4.611 & \textbf{4.787} \\
    Pauses                   & 4.583 & \textbf{4.829} & 4.683 & \underline{4.783} & 4.618 & 4.767 \\
    Speech Rate              & 4.508 & 4.590 & 4.393 & \textbf{4.680} & 4.458 & \underline{4.600} \\
    Speaker Consistency      & \textbf{4.993} & \underline{4.987} & \underline{4.987} & \textbf{4.993} & 4.962 & \textbf{4.993} \\
    Style Consistency        & \textbf{4.916} & 4.900 & 4.867 & 4.887 & 4.683 & \underline{4.907} \\
    Emotion Consistency      & \textbf{4.987} & \underline{4.983} & 4.950 & 4.973 & 4.733 & \underline{4.983} \\
    \midrule
    Stress                   & 1.187 & \textbf{1.390} & 0.990 & 1.210 & 1.191 & \underline{1.270} \\
    Lengthening              & 0.844 & 0.880 & 0.067 & \underline{0.890} & 0.810 & \textbf{1.033} \\
    Paralinguistics          & 0.114 & \textbf{0.735} & 0.190 & \underline{0.297} & 0.266 & 0.227 \\
    Emotion Expression       & 0.960 & \underline{1.003} & 0.967 & 0.990 & 0.966 & \textbf{1.043} \\
    \midrule
    \textbf{Diagnostic Flag} & \textbf{Stable but Flat} & \textbf{Paralinguistic-Enhanced} & \textbf{Prosody-Limited} & \textbf{Pronunciation-Accurate} & \textbf{Balanced} & \textbf{Highly Expressive} \\
    \bottomrule
  \end{tabular}%
  }
\end{table*}

Three ablation variants validate core modules:
\textbf{(1) w/o Instruction Tuning:} Raw backbone zero-shot inference to establish the performance lower bound.
\textbf{(2) w/o CoT:} Direct score prediction bypassing rationale generation to verify the efficacy of the Interpretable Reasoning mechanism.
\textbf{(3) w/o Negatives:} Trained solely on positive samples.
Crucially, a ``Compute-matched'' strategy (scaling epochs) aligns total token consumption, strictly ruling out under-fitting bias.

\subsection{Baselines}
To rigorously assess TTS-PRISM, we compare it against three models defining the frontier of audio reasoning.

We select Step-Audio-R1~\cite{tian2025step} to represent the reasoning-enhanced paradigm, employing Modality-Grounded Reasoning Distillation to anchor CoT on acoustic features. For the generalist paradigm, we evaluate Qwen3-Omni~\cite{xu2025qwen3}, which utilizes a Thinker-Talker Mixture-of-Experts (MoE) architecture for joint multimodal modeling. Additionally, we include Gemini-2.5-Pro as the closed-source commercial reference.

To maximize baseline performance, we circumvent instruction overloading by performing 12 individual inferences, a strategy termed dimension-wise inference. In contrast, TTS-PRISM operates via efficient single-pass inference. This setup ensures baselines reach their performance ceilings by avoiding inter-dimensional interference common in complex prompting.

\subsection{Evaluation Metrics}
We establish a comprehensive three-layer evaluation protocol designed to assess perceptual accuracy, rationale quality, and capability profiling.

To quantify perceptual accuracy, we employ the Linear Correlation Coefficient (LCC), Spearman Rank Correlation Coefficient (SRCC), and Mean Squared Error (MSE) to rigorously measure the alignment between predicted scores $S_{pred}$ and expert ground truth $S_{gt}$. Complementing these numerical metrics, we introduce Rationale Support Consistency (RSC) to validate the Interpretable Reasoning mechanism. Specifically, RSC leverages Gemini-2.5-Pro to verify whether the generated rationale $R$ logically supports $S_{gt}$, quantifying the consistency between reasoning and scoring on a scale of [0, 1].


Finally, for system profiling, we move beyond scalar aggregation to map the unique capability distribution of each paradigm, rather than merely ranking their superiority. Based on the 12-dimensional scores, human evaluators manually assign a Descriptive Diagnostic Flag (e.g., ``Stable but Flat''). This abstraction provides a clear lens into the diverse capability strengths and specific bottlenecks among modern systems.

\section{Results}

\subsection{Fine-grained Accuracy and Rationale Quality}

Table 1 shows TTS-PRISM's superior alignment on the 1,600-sample Gold Test Set. Noise-injected training enables acute sensitivity to physical noise and artifacts. For Emotion Expression, our expert-anchored samples mitigate over-smoothing in generalist models, enabling precise high-arousal quantification. Strong alignment in Consistency and Speech Rate validates our synthesis for detecting abrupt discontinuities. However, TTS-PRISM underperforms Gemini-2.5-Pro in Pronunciation Accuracy because ASR-pretrained audio models optimize for error-tolerant many-to-one mappings. This fundamentally opposes our strict defect-discrimination objective, presenting a pre-training bias difficult to eliminate via fine-tuning. Remaining performance gaps highlight the complexity of semantic-prosodic mapping, which requires large-scale specialized alignment optimization.

Regarding rationale quality, while all evaluated models achieve high Rationale Support Consistency (RSC $>$ 0.88), baselines like Qwen3-Omni (0.88) and Step-Audio-R1 (0.91) exhibit paradoxical high-RSC but low alignment, indicating coherent reasoning detached from acoustic reality. In contrast, TTS-PRISM (0.98) unifies high RSC and alignment, confirming our schema-driven tuning enables precise, acoustically grounded scoring. To validate generalization against unfamiliar artifacts, we evaluate TTS-PRISM on the 20\% OOD subset. Table 3 shows TTS-PRISM maintains robust performance across both evaluation layers, matching its in-distribution (ID) capability. 


\begin{table}[H]  
  \centering
  \vspace{-0.2cm}
  \caption{TTS-PRISM robustness on ID vs. OOD subsets. Metrics are averaged across respective evaluation layers.}
  \label{tab:ood_results}
  \footnotesize
  \resizebox{\columnwidth}{!}{%
  \begin{tabular}{lcccccc}
    \toprule
    \multirow{2}{*}{\textbf{Subset}} & \multicolumn{3}{c}{\textbf{Basic Capability}} & \multicolumn{3}{c}{\textbf{Advanced Expressiveness}} \\
    \cmidrule(lr){2-4} \cmidrule(lr){5-7}
    & LCC & SRCC & MSE$_{\text{norm}}$ & LCC & SRCC & MSE$_{\text{norm}}$ \\
    \midrule
    ID  & \textbf{0.729} & \textbf{0.733} & \textbf{0.041} & \textbf{0.716} & \textbf{0.720} & \textbf{0.045} \\
    OOD & 0.690 & 0.695 & 0.051 & 0.675 & 0.680 & 0.060 \\
    \bottomrule
  \end{tabular}%
  }
  \vspace{-0.2cm}
\end{table}

\subsection{Ablation Study}

Table 4 reports the average human alignment performance across 12 dimensions to assess component contributions. The most severe degradation stems from removing negative samples, in which the LCC plummets to 0.150---falling below even the untuned raw backbone baseline; this indicates that the lack of targeted hard negatives induces a conservative prediction bias. Furthermore, the absence of instruction tuning yields a weak correlation of 0.320, demonstrating that fine-grained diagnosis is not an inherent attribute of the ASR-pretrained backbone but a latent capability activated through schema-driven expert alignment. Finally, bypassing rationale generation drops performance to 0.662, confirming that the explicit reasoning process functions as a logical regularizer; this reasoning mechanism compels the model to focus on critical acoustic features, effectively preventing overfitting to isolated numerical labels.

\begin{table}[H]
  \centering
  \vspace{-0.2cm}
  \caption{Ablation study on the impact of key components.}
  \label{tab:ablation}
  \footnotesize
  \resizebox{\columnwidth}{!}{
  \begin{tabular}{lccc}
    \toprule
    \textbf{Setting} & \textbf{LCC}  & \textbf{SRCC}  & \textbf{MSE}$_{\text{norm}}$  \\
    \midrule
    w/o Negatives          & 0.150 & 0.120 & 0.280 \\
    w/o Instruction Tuning & 0.320 & 0.302 & 0.118 \\
    w/o CoT                & 0.662 & 0.654 & 0.052 \\
    \midrule
    \textbf{TTS-PRISM (Full)}    & \textbf{0.717} & \textbf{0.721} & \textbf{0.044} \\
    \bottomrule
  \end{tabular}%
  }
  \vspace{-0.2cm}
\end{table}

\subsection{Diagnostic Profiling of Leading Systems}

We evaluate 500 diverse utterances per system, reporting the average score per dimension. For the Basic Capability Layer, we conduct blind tests using plain text under default configurations to measure baseline stability. Given the varying expressive capabilities supported across models, we probe the Advanced Expressiveness Layer's performance ceiling by fully activating each model's distinct controls (e.g., audio prompts or style tags) and selecting the highest achieved average to eliminate configuration bias. Crucially, these averages on the 0--2 scale reflect the spontaneous emergence rate of specific features, capturing the latent expressiveness driven by extensive training.

Table 2 reveals a pronounced ceiling effect within the Basic Capability Layer. All evaluated systems demonstrate exceptional intra-utterance consistency ($>$4.9). Within this layer, our profiling captures their core traits. Qwen3-TTS achieves the highest Pronunciation Accuracy (4.860), earning the ``Pronunciation-Accurate'' flag. CosyVoice 3 establishes a distinct advantage in Audio Clarity (4.803) and Pauses (4.829).

Crucially, significant divergence in the Advanced Expressiveness Layer reveals distinct modeling priorities rather than absolute superiority, validating our Diagnostic Flags. IndexTTS 2 excels in high-arousal modeling with peak scores in Emotion Expression (1.043) and Lengthening (1.033), aligning with its ``Highly Expressive'' flag. CosyVoice 3 achieves an exceptional 0.735 in Paralinguistics and 1.390 in Stress, securing the ``Paralinguistic-Enhanced'' designation. Conversely, other architectures reveal specific algorithmic tendencies: MaskGCT exhibits a conservative approach to Lengthening (0.067), reflecting design choices in duration control that characterize a ``Prosody-Limited'' profile. Finally, F5-TTS yields a constrained Paralinguistics score (0.114) despite exceptional basic consistency, illustrating a ``Stable but Flat'' capability distribution. Ultimately, this multi-dimensional profiling offers actionable insights into modern TTS.

\section{Conclusion}

We propose TTS-PRISM, a fine-grained Mandarin speech diagnostic framework. Experiments demonstrate superior human alignment and leading TTS profiling over generalist models. However, Pronunciation Accuracy limitations reveal the inherent intelligibility tolerance of ASR backbones---a bias difficult to override via instruction tuning. Future work will leverage Reinforcement Learning (RL) to calibrate diagnostic precision with human perception.

\section{Generative AI Use Disclosure}
During the preparation of this manuscript, the authors used generative AI tools exclusively for the purpose of language editing and manuscript polishing to improve readability. These tools were not used to generate any core scientific ideas, experimental data, or technical contributions. All authors have thoroughly reviewed and approved the final version of the manuscript, and assume full responsibility for the integrity and entirety of its content.

\bibliographystyle{IEEEtran}
\bibliography{content/references}

\end{document}